\documentclass[11pt]{article}
\usepackage{acl2015}
\usepackage{times}
\usepackage{url}
\usepackage{latexsym}

\usepackage{caption} 
\usepackage{subcaption} 
\usepackage{multirow} 
\usepackage{amsmath} 
\usepackage{graphicx}
\usepackage{color} 
\newcommand{\imgExt}{eps}

\newcommand{\edit}[1]{{#1}} %
\newcommand{\hide}[1]{}
\newcommand{\unk}[1]{{\bf {\it {\underline{#1}}}}}
\newcommand{\unksym}{{\it \texttt{\underline{unk}}}}
\newcommand{\eossym}{$<$\texttt{eos}$>$}
\newcommand{\unkcopy}[1]{{\bf {\it {\texttt{\underline{unk}$_{#1}$ }}}}}
\newcommand{\unknull}{{\bf {\it {\texttt{\underline{unk}$_{\emptyset}$}}}}}
\newcommand{\unktext}[1]{{\bf {\it \texttt{\underline{unk}}$_{#1}$}}}
\newcommand{\pos}[1]{{\bf {\it {\texttt{p$_{#1}$}}}}}
\newcommand{\posnull}{{\bf {\it {\texttt{p$_{\emptyset}$}}}}}
\newcommand{\postext}[1]{{\bf {\it \texttt{p}$_{#1}$}}}
\newcommand{\unkpos}[1]{{\bf {\it {\texttt{\underline{unkpos}$_{#1}$}}}}}
\newcommand{\unkpostext}[1]{{\bf {\it \texttt{unkpos}$_{#1}$}}}

\newcommand{\bestbleu}{35.6}
\newcommand{\bestbleuunk}{37.5} 
\newcommand{\bestbleuunkwmt}{36.6} 
\newcommand{\bestunkimp}{2.8} 
\newcommand{\unkimp}{1.9} 
\newcommand{\imprare}{4.8} 

\title{Addressing the Rare Word Problem in \\ Neural Machine Translation}
\author{
Minh-Thang Luong$^\dagger$ \thanks{Work done while the authors were 
in Google.   $\dagger$ indicates equal contribution.} \\
Stanford \\
\texttt{lmthang@stanford.edu} \\
\AND
Ilya Sutskever$^\dagger$ \\
Google \\
\And
Quoc V. Le$^\dagger$ \\
Google \\
\texttt{\{ilyasu,qvl,vinyals\}@google.com} \\
\And
Oriol Vinyals \\
Google \\
\And
Wojciech Zaremba$^*$ \\
New York University \\
\texttt{woj.zaremba@gmail.com} \\
}

\date{}

\begin{document}
\maketitle
\begin{abstract}
Neural Machine Translation (NMT) is a new approach
to machine translation that has shown promising results that are comparable
to traditional approaches. 
A significant weakness in conventional NMT 
systems is their inability to correctly translate very rare words:  
end-to-end NMTs tend to have relatively small vocabularies with a single
\unktext{} symbol that represents every possible out-of-vocabulary (OOV) word. In
this paper, we propose and implement an effective technique to address this
problem. We train an NMT system on data that is augmented by the output of a word 
alignment algorithm, allowing the NMT system to emit, for each OOV word
in the target sentence, the position of its corresponding word in the source sentence.
This information is later utilized in a
post-processing step that translates every OOV word using a dictionary.  Our
experiments on the WMT'14 English to French translation task show that this 
method provides a substantial improvement of up to \bestunkimp{} BLEU points over an
equivalent NMT system that does not use this technique. 
With \bestbleuunk{} BLEU points, our NMT system is the first to surpass 
the best result achieved on a WMT'14 contest task.
\end{abstract}

\section{Introduction}
\label{sec:intro}

\begin{figure*}[tbh!]
\setlength{\unitlength}{1cm}
\begin{picture}(10, 2.7) 
\put(0,2){{\it en}: The \unk{ecotax} portico in \unk{Pont-de-Buis} , \ldots [truncated] \ldots , was taken down on Thursday morning}
\put(0,1){{\it fr}: \mbox{} Le \unk{portique} \unk{\'{e}cotaxe} de \unk{Pont-de-Buis} , \ldots [truncated] \ldots , a \'{e}t\'{e} \unk{d\'{e}mont\'{e}} jeudi matin}
\put(0,0){{\it nn}: Le \unksym{} de \unksym{} \`{a} \unksym{} , \ldots [truncated] \ldots , a \'{e}t\'{e} pris le jeudi matin}
\put(1.5,1.3){\line(2,1){1.2}} 
\put(2.8,1.3){\line(-2,1){1.2}} 
\put(4.8,1.3){\line(-1,2){0.3}} 
\put(10,1.3){\line(-1,1){0.6}} 
\put(10,1.3){\line(1,3){0.2}} 
\put(11,1.3){\line(3,2){0.9}} 
\put(11.9,1.3){\line(2,1){1.2}} 
\end{picture}
\caption{{\bf Example of the rare word problem} -- An English source sentence ({\it en}), a human translation to French ({\it fr}), and a translation produced by one of our neural network systems ({\it nn}) before handling OOV words. We highlight \unk{words} that are unknown to our model. 
The token \unksym{} indicates an OOV word. 
We also show a few important alignments between the pair of sentences. 
}
\label{f:sent_pair}
\end{figure*}

Neural Machine
Translation (NMT) is a novel approach to MT that has 
achieved promising results \cite{kal13,sutskever14,cho14,bog15,jean15}. 
An NMT system is a conceptually simple large neural network that 
reads the entire source sentence and produces an output translation one word at a time.
NMT systems are appealing because they use minimal domain knowledge which
makes them well-suited to any problem that can be formulated as mapping an input sequence 
to an output sequence \cite{sutskever14}.
In addition, the natural ability of neural networks to generalize implies that 
NMT systems will also generalize to novel word phrases and sentences that do not occur in the
training set. 
In addition, NMT systems potentially remove the need to store explicit phrase tables 
and language models which are used in conventional systems. 
Finally, the decoder of an NMT system is easy to implement, unlike the highly
intricate decoders used by phrase-based systems \cite{Koehn:2003:SMT}.

Despite these advantages, conventional NMT systems are incapable of translating rare 
words because they have a fixed modest-sized vocabulary\footnote{ Due to the computationally intensive nature of the softmax, NMT systems often limit 
their vocabularies to be the top 30K-80K most frequent words in each language. However, \newcite{jean15}
has very recently proposed an efficient approximation to the softmax that \edit{allows
for training NTMs with very large vocabularies. As discussed in Section~\ref{sec:nmt}, this technique is complementary to ours.}}
which forces them to use the \unksym{} symbol to 
represent the large number of out-of-vocabulary (OOV) words, as illustrated in Figure~\ref{f:sent_pair}.
Unsurprisingly, both \newcite{sutskever14} and \newcite{bog15} have
observed that sentences with many rare words tend to be translated much more poorly than sentences
containing mainly frequent words.
Standard phrase-based systems \cite{koehn2007moses,chiang07hiero,cer10phrasal,dyer10cdec}, 
on the other hand, do not suffer from the rare word 
problem to the same extent because they can support a much larger vocabulary, 
and because their use of explicit alignments
and phrase tables allows them to memorize the translations 
of even extremely rare words. 

Motivated by the strengths of standard phrase-based system, we
propose and implement a novel approach to address the rare word problem of NMTs.
Our approach annotates the training corpus with 
explicit alignment information that enables the NMT system to emit, for each OOV word, a
``pointer'' to its corresponding word in the source sentence. This
information is later utilized in a post-processing step that translates
the OOV words using a dictionary or with the identity translation, if no translation is found.

Our experiments confirm that this approach is effective. On the English to French WMT'14
translation task, this approach provides an improvement of
up to \bestunkimp{} (if the vocabulary is relatively small) 
BLEU points over an equivalent NMT system that does not use this technique.
Moreover, our system is the first NMT that outperforms the winner of a WMT'14 task.

\section{Neural Machine Translation}
\label{sec:nmt}

A neural machine translation system is any neural network that maps a source 
sentence, $s_1,\ldots,s_n$,
to a target sentence, $t_1,\ldots,t_m$, where all sentences are assumed to 
terminate with a special
``end-of-sentence'' token \eossym{}.  More concretely, an NMT system uses a neural 
network to parameterize the conditional distributions
\begin{equation}
p(t_j|t_{<j},s_{\leq n})
\end{equation}
for $1\leq j \leq m$.  By doing so, it becomes possible to 
compute and therefore maximize the log probability
of the target sentence given the source sentence
\begin{equation}
\log p(t|s) = \sum_{j=1}^m  \log p\left(t_j|t_{<j},s_{\leq n}\right)
\end{equation}
There are many ways to parameterize these conditional distributions.
For example, \newcite{kal13} used a combination of a
convolutional neural network and a recurrent neural network, \newcite{sutskever14} used a deep Long Short-Term Memory
(LSTM) model, \newcite{cho14} used an architecture similar to the LSTM, and
\newcite{bog15} used a more elaborate neural network
architecture that uses an attentional mechanism over the input sequence, 
similar to \newcite{graves13c} and \newcite{graves14}.  

In this work, we use the model of \newcite{sutskever14}, which 
uses a deep LSTM to encode the input sequence and a separate deep LSTM 
to output the translation. The encoder reads the 
source sentence, one word at a time, and produces
a large vector that represents the entire source sentence. 
The decoder is initialized with this vector
and generates a translation, one word at a time, 
until it emits the end-of-sentence symbol \eossym{}.

None the early work in neural machine translation systems has addressed the rare word problem,
but the recent work of \newcite{jean15} has tackled it with 
\edit{an efficient approximation to the softmax to accommodate for a very large vocabulary (500K words). However, even with a large vocabulary, the problem with rare words, e.g., names, numbers, etc., still persists, and \newcite{jean15} found that using techniques similar to ours are beneficial and complementary to their approach.}


\section{Rare Word Models}
\label{sec:rare}
Despite the relatively large amount of work done on pure neural machine translation systems, 
there has been no work addressing the OOV problem in NMT systems, 
with the notable exception of \newcite{jean15}'s work mentioned earlier. 

We propose to address the rare word problem by training the NMT system
to track the origins of the unknown words in the target sentences.  If
we knew the source word responsible for each unknown target word, we could introduce
a post-processing step that would replace each \unksym{} in the system's output
with a translation of its source word, using 
either a dictionary or the identity translation.  For example, in
Figure~\ref{f:sent_pair}, if the model knows that the second unknown token 
in the NMT (line {\it nn}) originates from the source
word \texttt{ecotax}, it can perform a word dictionary lookup to
replace that unknown token by \texttt{\'{e}cotaxe}. Similarly, an
identity translation of the source word \texttt{Pont-de-Buis} can be
applied to the third unknown token.

We present three annotation strategies that can easily be applied to any NMT system \cite{kal13,sutskever14,cho14}. 
We treat the NMT system as a black box and train it on a corpus annotated by one of the models below. 
First, the alignments are produced with an unsupervised aligner. 
Next, we use the alignment links to construct a word dictionary that will 
be used for the word translations in the post-processing step.\footnote{When a source word has multiple translations, we use the translation with the highest probability. These translation probabilities are estimated from the unsupervised alignment links. When constructing the dictionary from these alignment links, we add a word pair to the dictionary only if its alignment count exceeds 100.}
If a word does not appear in our dictionary, then we apply the identity translation.

The first few words of the sentence pair in Figure~\ref{f:sent_pair} (lines {\it en}
and {\it fr}) illustrate our models. 

\subsection{Copyable Model}
\label{subsec:copyable}
\begin{figure}
\resizebox{10.5cm}{!}{
\setlength{\unitlength}{1cm}
\begin{picture}(10, 1) 
\put(0.5,0.7){en: The \unkcopy{1} portico in \unkcopy{2} \ldots}
\put(0.5,0){fr: \mbox{} Le \unknull{} \unkcopy{1} de \unkcopy{2} \ldots}
\end{picture}
}
\caption{ {\bf Copyable Model} -- an annotated example with two 
types of unknown tokens: ``copyable'' \unktext{n} and null \unknull{}.}
\label{f:copyable}
\end{figure}

In this approach, we introduce multiple tokens to represent the various unknown words in the 
source and in the target language, as opposed to using only one \unksym{} token. 
We annotate the OOV words in the source sentence
with \unktext{1}, \unktext{2}, \unktext{3}, in that order,
while assigning repeating unknown words identical tokens. 
The annotation of the unknown words in the target language is slightly more elaborate: (a) each 
unknown target word that is aligned to an unknown source word
is assigned the same unknown token (hence, the ``copy'' model) and 
(b) an unknown target word that has no 
alignment or that is aligned with a known word uses the special null token \unknull{}. 
See Figure~\ref{f:copyable} for an example.  This annotation enables us to 
translate every non-null unknown token.

\subsection{Positional All Model (PosAll)}
The copyable model is limited by its inability to translate unknown 
target words that are aligned to \emph{known} words in the source sentence, such as the pair of 
words, ``portico'' and ``portique'', in our running example. 
The former word is known on the source sentence; whereas latter is not, so it is labelled with \unknull{}.
This happens often since the source vocabularies of our models tend to be much 
larger than the target vocabulary since a large source vocabulary is cheap.
This limitation motivated us to develop an annotation model that includes the complete 
alignments between the source and the target sentences, which is straightforward to obtain
 since the complete alignments are available at training time.  

Specifically, we return to using only a single universal \unksym{} token. 
However, on the target side, 
we insert a positional token \postext{d} after every word. Here, $d$ indicates a relative position 
($d=-7,\ldots,-1,0,1,\ldots,7$) to denote that a target word at position $j$ is aligned 
to a source word 
at position $i=j-d$. Aligned words that are too far apart are considered unaligned, and 
unaligned words rae annotated
with a null token \postext{n}. Our annotation is illustrated in 
Figure~\ref{f:pos_all}.

\begin{figure}
\resizebox{10.5cm}{!}{
\setlength{\unitlength}{1cm}
\begin{picture}(10, 1) 
\put(0,0.7){en: The \unksym{} portico in \unksym{} \ldots} 
\put(0,0){fr: \mbox{} Le \pos{0} \unksym{} \pos{-1} \unksym{} \pos{1} de \posnull{} \unksym{} \pos{-1} \ldots}
\end{picture}
}
\caption{ {\bf Positional All Model} -- an example of the PosAll model. Each word is followed by the relative positional tokens \postext{d} or the null token \posnull{}. }
\label{f:pos_all}
\end{figure}

\subsection{Positional Unknown Model (PosUnk)}

The main weakness of the PosAll model is that it doubles the length of the target sentence. This
makes learning more difficult and slows the speed of parameter updates by a factor of two.
However, given that our post-processing step is concerned only with the alignments of the unknown words,
so it is more sensible to only annotate the unknown words. 
This motivates our {\it positional unknown} model which uses \unkpostext{d} 
tokens (for $d$ in $-7,\ldots,7$ or $\emptyset$) to simultaneously 
denote (a) the fact that a word is unknown and (b) its relative position $d$ with respect to its aligned source word. 
Like the PosAll model, we use the symbol \unkpos{\emptyset} for unknown target words that do not have an alignment. 
We use the universal \unksym{} for all unknown tokens in the source language. See Figure~\ref{f:pos_unk} for an annotated example.

\begin{figure}[tbh!]
\resizebox{10.5cm}{!}{
\setlength{\unitlength}{1cm}
\begin{picture}(10, 1) 
\put(0,0.7){en: The \unksym{} portico in \unksym{} \ldots} 
\put(0,0){fr: \mbox{} Le \unkpos{1} \unkpos{-1} de \unkpos{1} \ldots}
\end{picture}
}
\caption{ {\bf Positional Unknown Model} -- 
an example of the PosUnk model: only aligned unknown words are annotated with the \unkpostext{d} tokens.}
\label{f:pos_unk}
\end{figure}

It is possible that despite its slower speed, the PosAll model will learn better alignments because 
it is trained on many more examples of words and their alignments. 
However, we show that this is not the case (see $\S$\ref{subsec:rare_model_compare}).


\section{Experiments}
\label{sec:exp}
We evaluate the effectiveness of our OOV models on the WMT'14 English-to-French translation 
task.
Translation quality is measured with the BLEU metric \cite{Papineni02bleu} on the newstest2014 test set (which has 3003 sentences).

\subsection{Training Data}

To be comparable with the results reported by previous work on neural machine translation systems
\cite{sutskever14,cho14,bog15}, we train our models on 
the same training data of 12M parallel sentences (348M French and 304M English words), obtained from \cite{wmt14_en_fr}. 
The 12M subset was selected 
from the full WMT'14 parallel corpora using the method proposed in \newcite{Axelrod:2011:DAV}.

Due to the computationally intensive nature of the naive softmax,
we limit the French vocabulary (the {\it target} language) 
to the either the 40K or the 80K most frequent French words. On the {\it source} side, 
we can afford a much larger vocabulary, so we use the 200K most frequent English words. 
The model treats all other words as unknowns.\footnote{When the French vocabulary has 40K words, there are
on average 1.33 unknown words per sentence on the target side of the test set.}

We annotate our training data using the three schemes described in the previous section. The alignment 
is computed with the Berkeley aligner \cite{liang06alignment} using its default settings.
We discard sentence pairs in which the source or the target sentence exceed 100 tokens.

\subsection{Training Details}
\label{subsec:train_details}
Our training procedure and hyperparameter choices are similar to those used by
\newcite{sutskever14}. In more details, we train multi-layer deep LSTMs, each of which has 
1000 cells, with 1000 dimensional embeddings. Like \newcite{sutskever14}, 
we reverse the words in the source sentences which 
has been shown to improve LSTM memory utilization and results in better translations of long sentences. 
Our hyperparameters can be summarized as follows: (a) the parameters are initialized uniformly  
in [-0.08, 0.08] for 4-layer models and [-0.06, 0.06] for 6-layer models, (b) SGD has a fixed learning rate of 0.7, (c) we train for 8 epochs (after
5 epochs, we begin to halve the learning rate every 0.5 epoch), (d) the size of the mini-batch is 128, 
and (e) we rescale the normalized gradient to ensure that its norm does not exceed 5 \cite{razvan}.

We also follow the GPU parallelization scheme proposed in \cite{sutskever14}, allowing us to  
reach a training speed of 5.4K words per second to train a depth-6 model with 200K source and 80K target vocabularies  
; whereas \newcite{sutskever14} achieved 6.3K words per 
second for a depth-4 models with 80K source and target vocabularies.
Training takes about 10-14 days on an 8-GPU machine.

\begin{table*}[tbh!]
\centering
\begin{tabular}{l|c|c|l}
\bf{System} & \bf{Vocab} & {\bf Corpus} & \bf{BLEU}\\
  \hline
State of the art in WMT'14 \cite{durrani-EtAl:2014:W14-33} & All & 36M & {\bf 37.0}\\
  \hline
\multicolumn{4}{c}{{\it Standard MT + neural components}}\\
  \hline
\newcite{wmt14_en_fr} -- neural language model & All & 12M & 33.3\\ %
\newcite{cho14}-- phrase table neural features & All & 12M & 34.5\\ %
\newcite{sutskever14} -- 5 LSTMs, reranking 1000-best lists & All & 12M & 36.5\\ %
  \hline
\multicolumn{4}{c}{{\it Existing end-to-end NMT systems}}\\
  \hline
\newcite{bog15} -- single gated RNN with search & 30K & 12M & 28.5\\
\newcite{sutskever14} -- 5 LSTMs & 80K & 12M & 34.8\\
\newcite{jean15} -- 8 gated RNNs with search + UNK replacement & 500K & 12M & 37.2\\
  \hline
\multicolumn{4}{c}{{\it Our end-to-end NMT systems }}\\
  \hline
Single LSTM with 4 layers  & 40K & 12M & 29.5\\ 
Single LSTM with 4 layers + PosUnk & 40K & 12M & 31.8 (+2.3) \\
Single LSTM with 6 layers & 40K & 12M & 30.4\\ 
Single LSTM with 6 layers + PosUnk & 40K & 12M & 32.7 (+2.3) \\
Ensemble of 8 LSTMs & 40K & 12M & 34.1 \\
Ensemble of 8 LSTMs + PosUnk & 40K & 12M & 36.9 (+2.8)\\
  \hline
Single LSTM with 6 layers & 80K & 36M & 31.5\\ 
Single LSTM with 6 layers + PosUnk & 80K & 36M & 33.1 (+1.6) \\
Ensemble of 8 LSTMs & 80K & 36M & \bestbleu{}\\
Ensemble of 8 LSTMs + PosUnk & 80K & 36M & {\bf \bestbleuunk{} (+\unkimp{})}\\
\end{tabular}
\caption{{\bf Tokenized BLEU on newstest2014} --  Translation results of various systems which differ in terms of: (a) the architecture, (b) the size of the vocabulary used, and (c) the training corpus, either using the full WMT'14 corpus of 36M sentence pairs or a subset of it with 12M pairs. 
We highlight the performance of our best system in bolded text and state the improvements obtained by our technique of handling rare words (namely, the PosUnk model). Notice that, \edit{for a given vocabulary size}, the more accurate systems achieve a greater improvement from the post-processing step.  This is the case because the more accurate models are able to pin-point the origin of an unknown
word with greater accuracy, making the post-processing more useful.
}
\label{t:results}
\end{table*}

\subsection{A note on BLEU scores}
We report BLEU scores based on both: (a) {\it detokenized} translations, i.e., WMT'14 style, to be comparable with results reported on the WMT website\footnote{\url{http://matrix.statmt.org/matrix}} and (b) {\it tokenized translations}, so as to be consistent with previous work \cite{cho14,bog15,wmt14_en_fr,sutskever14,jean15}.\footnote{The \texttt{tokenizer.perl} and \texttt{multi-bleu.pl} scripts are used to tokenize and score translations.}

The existing WMT'14 state-of-the-art system \cite{durrani-EtAl:2014:W14-33} achieves a detokenized BLEU score of 35.8 on 
the newstest2014 test set for English to French language pair (see Table~\ref{t:results_wmt}). In terms of the tokenized BLEU, its performance is 37.0 points (see Table~\ref{t:results}).
\begin{table}[tbh!]
\centering
\begin{tabular}{l|c}
\bf{System} & \bf{BLEU}\\
  \hline
Existing SOTA \cite{durrani-EtAl:2014:W14-33} & 35.8\\
  \hline
Ensemble of 8 LSTMs + PosUnk  & {\bf \bestbleuunkwmt{}}\\
\end{tabular}
\caption{{\bf Detokenized BLEU on newstest2014} -- translation results of the existing state-of-the-art system and our best system.}
\label{t:results_wmt}
\end{table}

\subsection{Main Results}
We compare our systems to others,including the current state-of-the-art MT system \cite{durrani-EtAl:2014:W14-33},
recent end-to-end neural systems, as well as phrase-based baselines with neural components.

The results shown in Table~\ref{t:results} demonstrate that our unknown word translation technique (in particular, the PosUnk model) significantly improves the translation quality for both the individual (non-ensemble) LSTM models and the ensemble models.\footnote{
For the 40K-vocabulary ensemble, we combine 5 models with 4 layers and 3 models with 6 layers. For the 80K-vocabulary ensemble, we combine 3 models with 4 layers and 5 models with 6 layers. Two of the depth-6 models are regularized with dropout, similar to \newcite{zaremba15} with the dropout probability set to 0.2.} 
For 40K-word vocabularies, the performance gains are in the range of 2.3-2.8 BLEU points. With larger vocabularies (80K), the performance gains are diminished, but our technique can still provide a nontrivial gains of 1.6-1.9 BLEU points. 

It is interesting to observe that our approach is more useful for ensemble models as compared to the individual ones. 
This is because the usefulness of the PosUnk model directly
depends on the ability of the NMT to correctly locate, for a given OOV target word, its corresponding word in the source sentence.  An ensemble of large models identifies these source words with greater accuracy.  This is why for the same vocabulary size, better models obtain a greater performance gain our post-processing step. 
e
Except for the very recent work of \newcite{jean15} that employs a similar unknown treatment strategy\footnote{Their unknown replacement method and ours both track the locations of target unknown words and use a word dictionary to post-process the translation. However, the mechanism used to achieve the ``tracking'' behavior is different. \newcite{jean15}'s uses the attentional mechanism to track the origins of all target words, not just the unknown ones. In contrast, we only focus on tracking unknown words using unsupervised alignments. Our method can be easily applied to any sequence-to-sequence models since we treat any model as a blackbox and manipulate only at the input and output levels.} as ours, our best result of \bestbleuunk{} BLEU outperforms all other NMT systems by a 
arge margin, and 
more importanly, our system has established a new record on the WMT'14 English to French translation.


\section{Analysis}
\label{sec:analysis}

We analyze and quantify the improvement obtained by our rare word translation approach and provide a detailed 
comparison of the different rare word techniques proposed in Section~\ref{sec:rare}. We also examine the effect of 
depth on the LSTM architectures and demonstrate a strong correlation between perplexities and BLEU scores. We also highlight 
a few translation examples where our models succeed in correctly translating OOV words, and present 
several failures.

\subsection{Rare Word Analysis}
To analyze the effect of rare words on translation quality, 
we follow Sutskever et al.~\cite{sutskever14} and sort sentences in 
newstest2014 by the average inverse frequency of their words. 
We split the test sentences into groups where the sentences within each group have a comparable number of rare words
 and evaluate each group independently. We evaluate our 
systems before and after translating the OOV words and compare with 
the standard MT systems -- we use the best system from the WMT'14 contest \cite{durrani-EtAl:2014:W14-33},
and neural MT systems -- we use the ensemble systems described in \cite{sutskever14} and Section~\ref{sec:exp}.

\begin{figure}
\centering
\includegraphics[width=0.5\textwidth, clip=true, trim= 95 0 230 0]{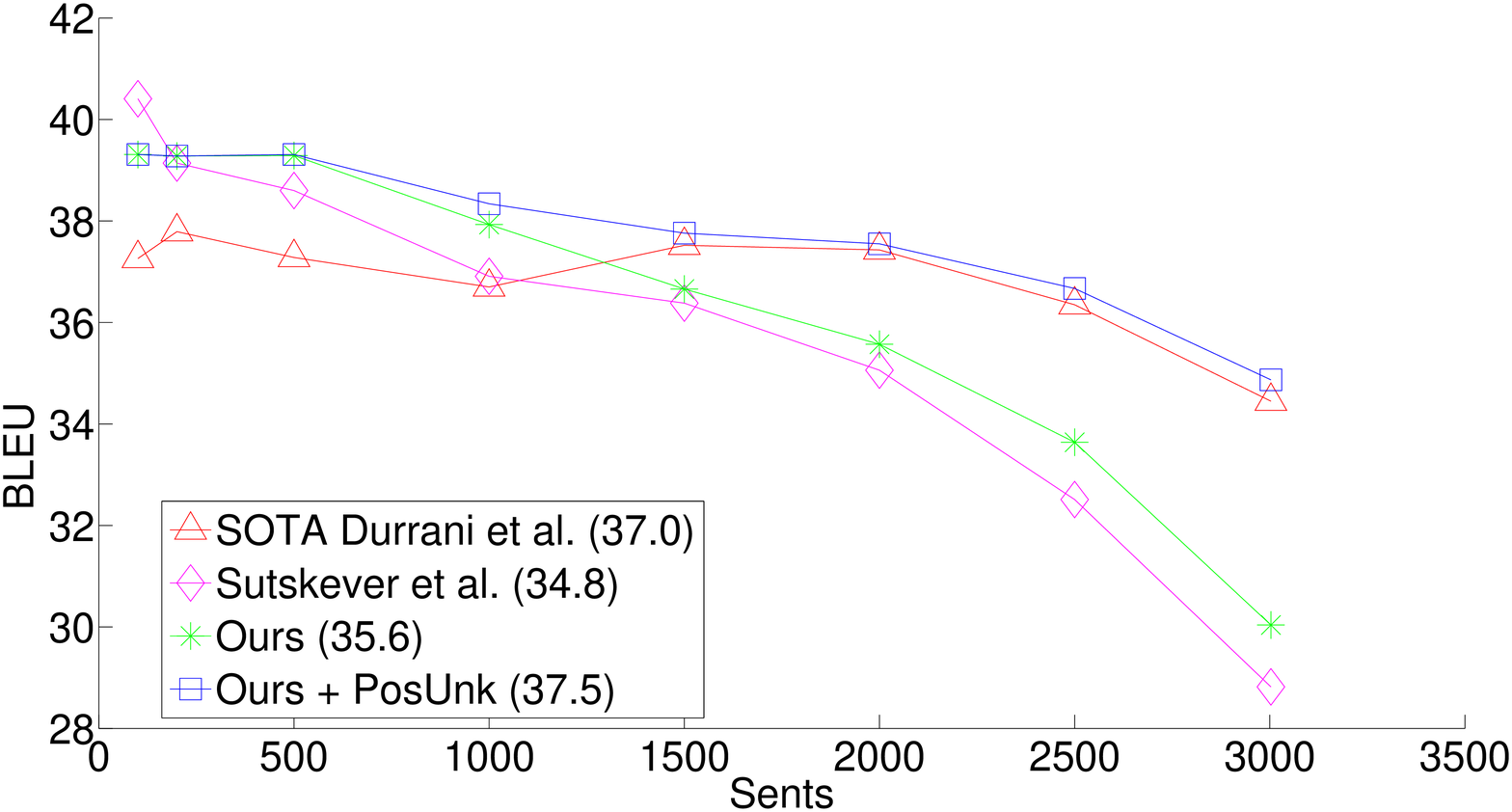} 
\caption{{\bf Rare word translation} -- 
On the x-axis, we order newstest2014 sentences by their {\it average frequency rank} and divide the sentences into groups 
of sentences with a comparable prevalence of rare words. 
We compute the BLEU score of each group independently.} 
\label{f:rare}
\end{figure}

Rare word translation is challenging for neural machine translation systems as
shown in Figure~\ref{f:rare}. Specifically, the translation quality of our
model before applying the postprocessing step is shown by the green curve, and the current
best NMT system \cite{sutskever14} is the purple curve. While \cite{sutskever14}
produces better translations for sentences with frequent words (the left part of the
graph), they are worse than best system (red curve)
on sentences with many rare words (the right side of the graph). When applying our
unknown word translation technique (purple curve), we
significantly improve the translation quality of our NMT: 
for the last group of 500 sentences which have the greatest proportion of 
OOV words in the test set, we increase the BLEU score of our system by 
\imprare{} BLEU points. Overall, our rare word translation model 
interpolates between the SOTA system and the system
of \newcite{sutskever14},  which allows us to outperform the winning entry of WMT'14
on sentences that consist predominantly of frequent words and approach its performance on sentences
with many OOV words.

\subsection{Rare Word Models}
\label{subsec:rare_model_compare}

We examine the effect of the different rare word models presented in
Section~\ref{sec:rare}, namely: (a) {\it Copyable} -- which aligns the unknown
words on both the input and the target side by learning to copy indices, (b) the Positional All
({\it PosAll}) -- which predicts the aligned source positions for every target
word, and (c) the Positional Unknown ({\it PosUnk}) -- which predicts the aligned
source positions for only the unknown target words.\footnote{In this section and in section~\ref{subsec:effects},
all models are trained on the unreversed sentences, and we use the following hyperparameters: 
we initialize the parameters uniformly in [-0.1, 0.1], the learning rate is 1, the maximal gradient norm is 1, 
with a source vocabulary of 90k words, and a target vocabulary of 40k (see Section~\ref{subsec:train_details} for more details).
While these LSTMs do not achieve the best possible performance, it is still useful to analyze them.}
It is also interesting to measure the improvement obtained when no alignment information is used during training.
As such, we include a baseline model with no alignment knowledge ({\it NoAlign}) in which we simply assume that the $i^{\textrm{th}}$ unknown word on the target
sentence is aligned to the $i^{\textrm{th}}$ unknown word in the source sentence.

\begin{figure}
\centering
\includegraphics[width=0.5\textwidth, clip=true, trim= 105 0 140 0]{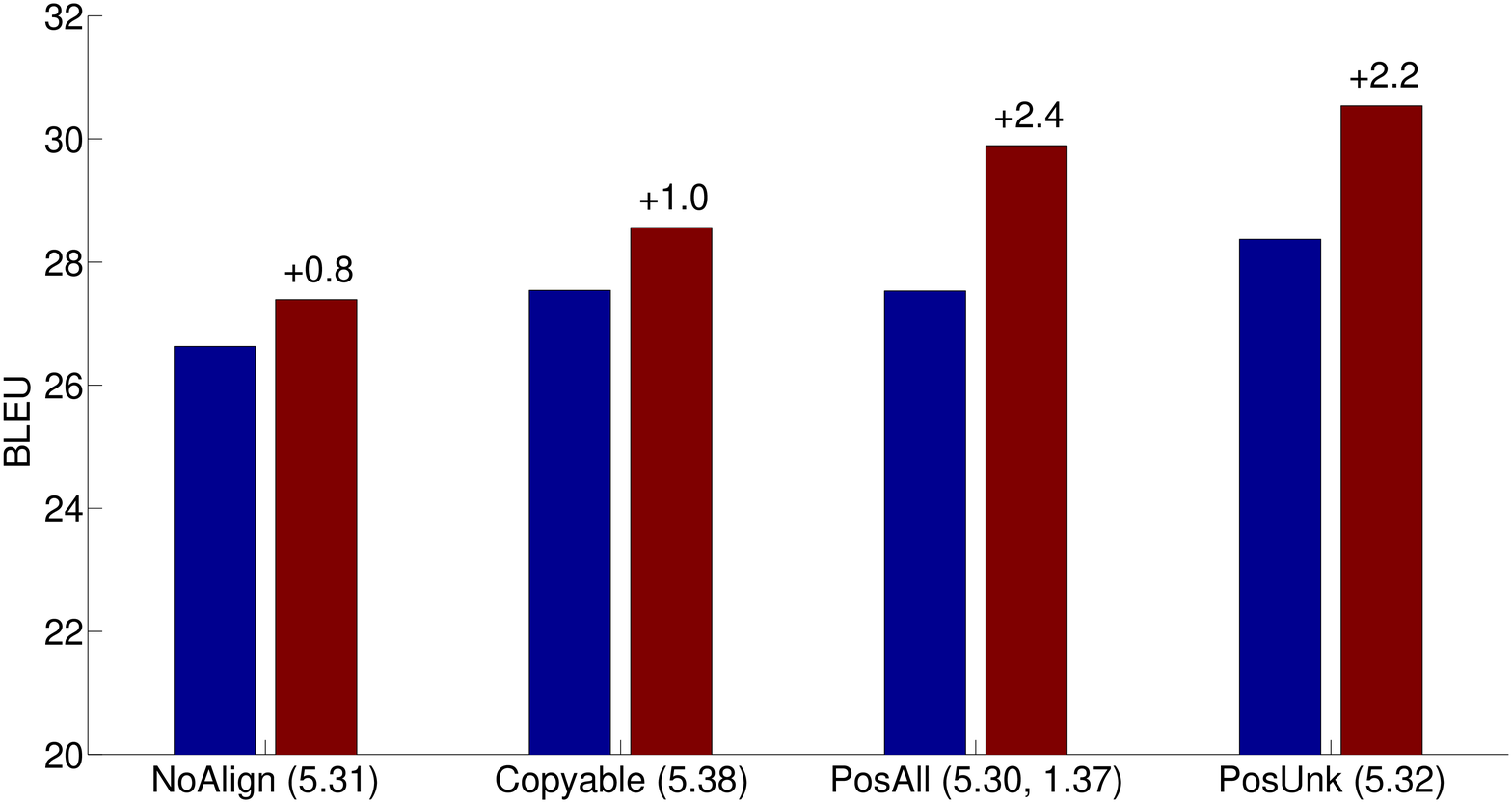} 
\caption{{\bf Rare word models} -- 
translation performance of 6-layer LSTMs:
a model that uses no alignment ({\it NoAlign}) 
and the other rare word models ({\it Copyable, PosAll, PosUnk}). 
For each model, we show results before ({\it left}) and after ({\it right}) the rare word translation as well as the perplexity (in parentheses).
For {\it PosAll}, we report the perplexities of predicting the words and the positions.} 
\label{f:compare}
\end{figure}

From the results in Figure~\ref{f:compare}, a simple monotone
alignment assumption for the {\it NoAlign} model yields a modest gain of
0.8 BLEU points. If we train the model to predict the alignment, then the {\it Copyable} model
offers a slightly better gain of 1.0 BLEU. Note, however, that English
and French have similar word order structure, so it would be
interesting to experiment with other language pairs, such as English and
Chinese, in which the word order is not as monotonic. These harder language pairs 
potentially imply a smaller gain for the NoAlign model and a larger
gain for the Copyable model. 
We leave it for future work.

\begin{figure}[tbh!]
\centering
\includegraphics[scale=0.18, clip=true, trim= 0 0 0 0]{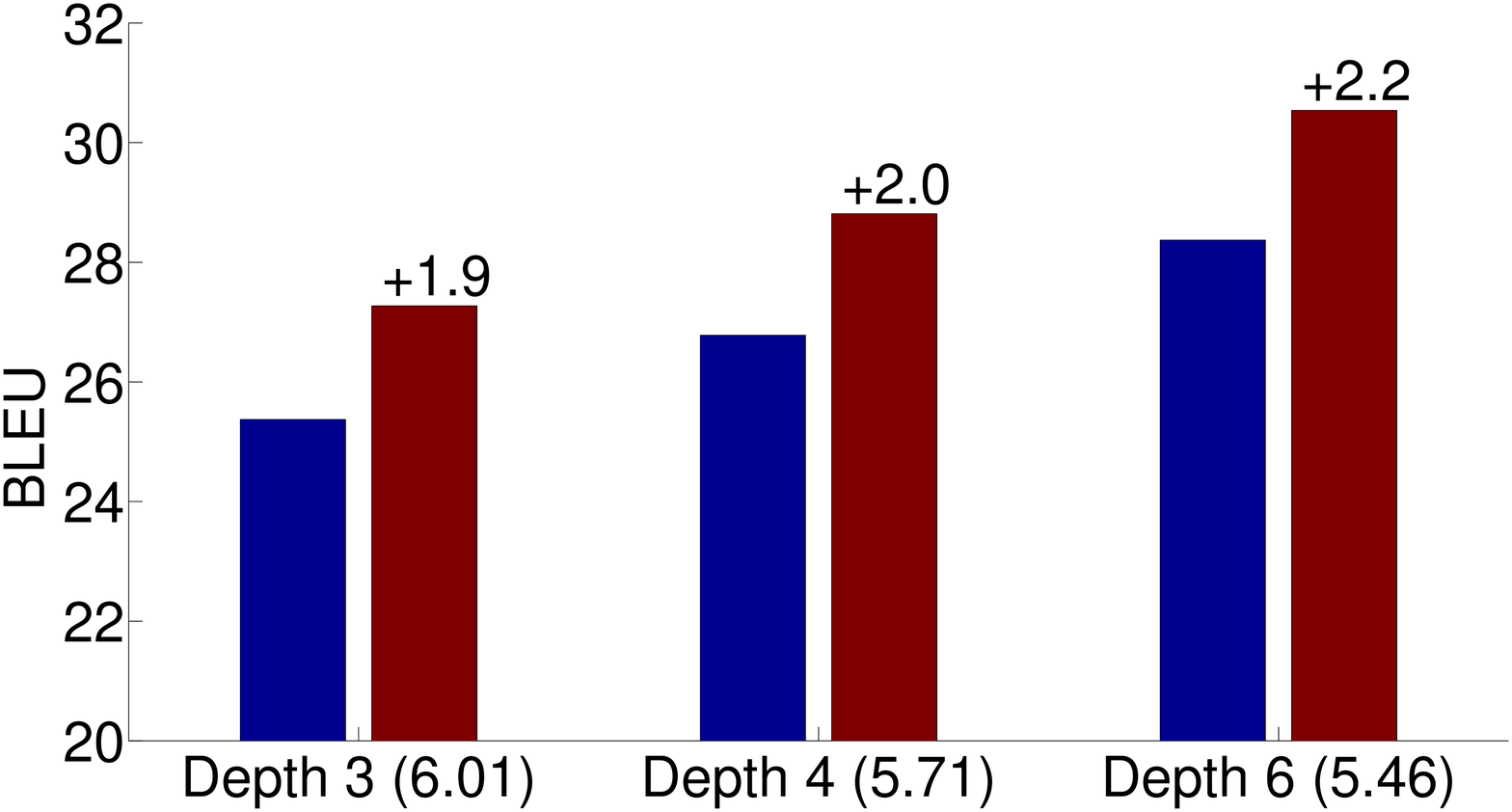} 
\caption{{\bf Effect of depths} -- BLEU scores achieved by {\it PosUnk} models of various depths (3, 4, and 6) before and after the rare word translation. 
 Notice that the PosUnk model is more useful on more accurate models. }
\label{f:depth}
\end{figure}

The positional models ({\it PosAll} and {\it PosUnk}) 
improve translation performance by more than 2 BLEU points. 
This proves that the limitation of the copyable model, which forces
it to align each unknown output word with an unknown input word, is considerable.  
In contrast, the positional models can align the unknown target words with any source word,
and as a result, post-processing has a much stronger effect. 
The PosUnk model achieves better translation results than
the PosAll model which suggests that it is easier to train the LSTM on shorter sequences. 

\subsection{Other Effects}
\label{subsec:effects}
{\bf Deep LSTM architecture} --  We compare PosUnk models trained with different number of layers (3, 4, and 6). 
We observe that the gain obtained by the PosUnk model increases in tandem with the overall accuracy of the model, which is consistent 
with the idea that larger models can point to the appropriate source word more accurately.
Additionally, we observe that on average, each extra LSTM layer provides roughly 1.0 BLEU point improvement as demonstrated in Figure~\ref{f:depth}. 

\begin{figure}[tbh!]
\centering
\includegraphics[width=0.5\textwidth, clip=true, trim= 0 0 0 0]{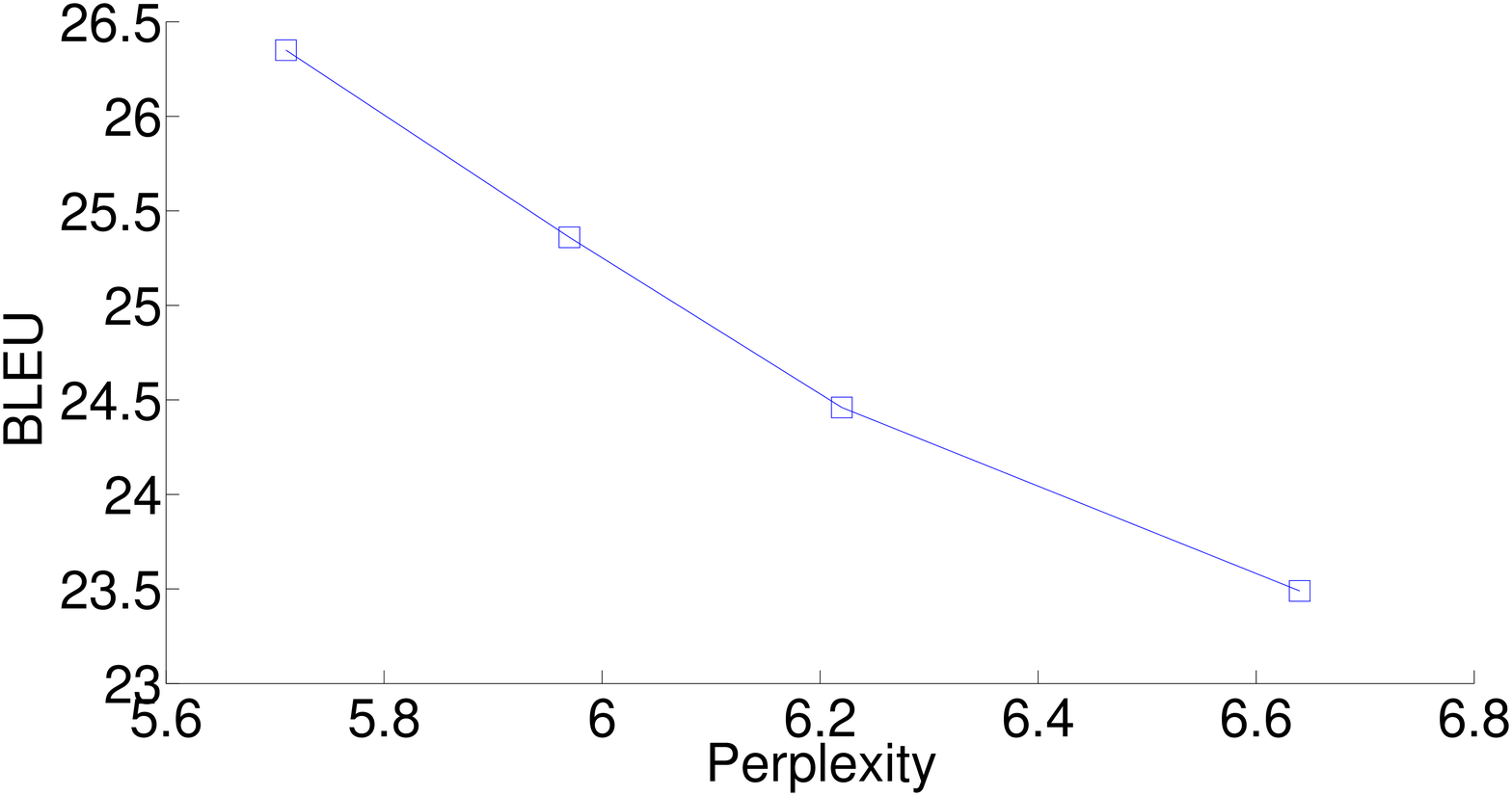} 
\caption{{\bf Perplexity vs. BLEU} -- we show the correlation by evaluating an LSTM model with 4 layers at various stages of training. 
} 
\label{f:cor}
\end{figure}

\begin{table*}[tbh!]
\centering
\resizebox{13.5cm}{!}{
\begin{tabular}{c|p{12cm}}
\bf{} & \bf{Sentences}\\
  \hline
src & An additional {\it 2600} operations including {\it orthopedic} and {\it cataract} surgery will help clear a backlog .\\
  \hline
trans & En outre , \unkpos{1} op{\'e}rations suppl{\'e}mentaires , dont la chirurgie \unkpos{5} et la \unkpos{6} , permettront de r{\'e}sorber l' arri{\'e}r{\'e} .\\
  \hline
+unk & En outre , {\it 2600} op{\'e}rations suppl{\'e}mentaires , dont la chirurgie {\it orthop{\'e}diques} et la {\it cataracte} , permettront de r{\'e}sorber l' arri{\'e}r{\'e} .\\
  \hline
tgt & 2600 op{\'e}rations suppl{\'e}mentaires , notamment dans le domaine de la chirurgie orthop{\'e}dique et de la cataracte , aideront {\`a} rattraper le retard .\\
  \hline
  \hline
src & This {\it trader} , Richard {\it Usher} , left RBS in {\it 2010} and is understand to have be given leave from his current position as European head of forex spot trading at {\it JPMorgan} .\\
  \hline
trans & Ce \unkpos{0} , Richard \unkpos{0} , a quitt{\'e} \unkpos{1} en 2010 et a compris qu' il est autoris{\'e} {\`a} quitter son poste actuel en tant que leader europ{\'e}en du march{\'e} des points de vente au \unkpos{5} .\\
  \hline
+unk & Ce {\it n\'{e}gociateur} , Richard {\it Usher} , a quitt{\'e} RBS en {\it 2010} et a compris qu' il est autoris{\'e} {\`a} quitter son poste actuel en tant que leader europ{\'e}en du march{\'e} des points de vente au {\it JPMorgan} .\\
  \hline
tgt & Ce trader , Richard Usher , a quitt{\'e} RBS en 2010 et aurait {\'e}t{\'e} mis suspendu de son poste de responsable europ{\'e}en du trading au comptant pour les devises chez JPMorgan \\
  \hline
  \hline
src & But concerns have grown after Mr {\it Mazanga} was quoted as saying {\it Renamo} {\it was} abandoning the 1992 peace accord .\\
  \hline
trans & Mais les inqui{\'e}tudes se sont accrues apr{\`e}s que M. \unkpos{3} a d{\'e}clar{\'e} que la \unkpos{3} \unkpos{3} l' accord de paix de 1992 .\\
  \hline
+unk & Mais les inqui{\'e}tudes se sont accrues apr{\`e}s que M. {\it Mazanga} a d{\'e}clar{\'e} que la {\it Renamo} {\it {\'e}tait} l' accord de paix de 1992 .\\
  \hline
tgt & Mais l' inqui{\'e}tude a grandi apr{\`e}s que M. Mazanga a d{\'e}clar{\'e} que la Renamo abandonnait l' accord de paix de 1992 .\\
\end{tabular}
}
\caption{{\bf Sample translations} -- the table shows the source ({\it src}) and the translations of our best model before ({\it trans}) and after ({\it +unk}) unknown word translations. We also show the human translations ({\it tgt}) and italicize words that are involved in the unknown word translation process.}
\label{t:sample}
\end{table*}

{\bf Perplexity and BLEU} -- Lastly, we find it interesting to observe a strong correlation 
between the perplexity (our training objective) and the translation quality as measured by BLEU. 
Figure~\ref{f:cor} shows the performance of a 4-layer LSTM, in which we compute both perplexity and 
BLEU scores at different points during training. We find that on average, a reduction of 0.5 perplexity 
gives us roughly 1.0 BLEU point improvement.

\subsection{Sample Translations}
We present three sample translations of our best system
(with \bestbleuunk{} BLEU) in Table~\ref{t:sample}. In our  first example,
the model translates all the
unknown words correctly: {\it 2600}, {\it orthop{\'e}diques}, and {\it
cataracte}. It is interesting to observe that the model can accurately predict
an alignment of distances of 5 and 6 words. The second
example highlights the fact that our model can translate long
sentences reasonably well and that it was able to
correctly translate the unknown word for {\it JPMorgan} at the very far end of
the source sentence. Lastly, our examples also reveal several
penalties incurred by our model: (a) incorrect entries in the word dictionary, as with {\it n\'{e}gociateur} vs. {\it trader} in the second example, 
and (b) incorrect alignment prediction, such as when 
 \unkpostext{3} is incorrectly aligned
with the source word {\it was} and not with {\it abandoning}, which resulted in an
incorrect translation in the third sentence.


\section{Conclusion}
\label{sec:conclude}

We have shown that a simple alignment-based technique can mitigate and even
overcome one of the main weaknesses of current NMT systems, which is
their inability to translate words that are not in their vocabulary.  
A key advantage of our technique is the fact that it is applicable to any NMT system and not
only to the deep LSTM model of \newcite{sutskever14}.  A technique
like ours is likely necessary if an NMT system is to achieve state-of-the-art performance
on machine translation.

We have demonstrated empirically that on the WMT'14 English-French translation task, our technique yields a 
consistent and substantial improvement of up to \bestunkimp{} BLEU points over various NMT systems of different architectures. 
Most importantly, with \bestbleuunk{} BLEU points, we have established the first NMT system that outperformed 
the best MT system on a WMT'14 contest dataset.

\section*{Acknowledgments}
We thank members of the Google Brain team for thoughtful discussions and insights. The first author especially thanks Chris Manning and the Stanford NLP group for helpful comments on the early drafts of the paper. Lastly, we thank the annonymous reviewers for their valuable feedback.
\bibliography{acl2015}
\bibliographystyle{acl2015} 

\end{document}